\documentclass{article}

\usepackage{arxiv}
\usepackage[utf8]{inputenc} 
\usepackage[T1]{fontenc}    
\usepackage{hyperref}       
\usepackage{url}            
\usepackage{booktabs}       
\usepackage{amsfonts}       
\usepackage{nicefrac}       
\usepackage{microtype}      
\usepackage{lipsum}		    
\usepackage{graphicx}
\usepackage{doi}
\usepackage{longtable}
\usepackage{multicol}
\usepackage{dblfloatfix}
\usepackage{caption}
\usepackage{lscape}

\title{A Topic Modeling Approach to Classifying Open Street Map 
Health Clinics and Schools in Sub-Saharan Africa}

\author{ 
    Joshua W.~Anderson \\
	Chapman University\\
	\texttt{joshanderson@chapman.edu} \\
	\And
    Luis Iñaki Alberro Encinas\\
    World Bank \\
    \texttt{lalberroencinas@worldbank.org} \\
    \And 
    Tina George Karippacheril\\
    World Bank\\
    \texttt{tgeorge1@worldbank.org}\\
    \And
	Jonathan Hersh \\
	Chapman University\\
	\texttt{hersh@chapman.edu} \\
    \And
    Cadence Stringer\\
    Chapman University\\
    \texttt{cstringer@chapman.edu}
}

\renewcommand{\shorttitle}{Topic Modeling OSM
Health Clinics and Schools in Sub-Saharan Africa}

\hypersetup{
pdftitle={A Topic Modeling Approach to Classifying Open Street Map 
Health Clinics and Schools in Sub-Saharan Africa},
pdfauthor={Joshua W. Anderson, Luis Encinas, Tina Karippacheril, Jonathan Hersh, Cadence Stringer},
pdfkeywords={unsupervised learning, unstructured data, development, social protection, open street map},
}

\begin{document}

\maketitle

\begin{abstract}
Data deprivation, or the lack of easily available and actionable information on the well-being of individuals, is a significant challenge for the developing world and an impediment to the design and operationalization of policies intended to alleviate poverty. In this paper we explore the suitability of data derived from OpenStreetMap to proxy for the location of two crucial public services: schools and health clinics. Thanks to the efforts of thousands of digital humanitarians, online mapping repositories such as OpenStreetMap contain millions of records on buildings and other structures, delineating both their location and often their use. Unfortunately much of this data is locked in complex, unstructured text rendering it seemingly unsuitable for classifying schools or clinics. We apply a scalable, unsupervised learning method to unlabeled OpenStreetMap building data to extract the location of schools and health clinics in ten countries in Africa. We find the topic modeling approach greatly improves performance versus reliance on structured keys alone. We validate our results by comparing schools and clinics identified by our OSM method versus those identified by the WHO, and describe OSM coverage gaps more broadly.
\end{abstract}

\section{Introduction}

In the wake of the COVID-19 pandemic, the World Bank’s 2020 Global Economic Prospects forecasts a baseline global GDP contraction of 5.2 percent, making it the deepest global recession in decades. Between 71 to 100 million people are expected to be pushed into extreme poverty, almost half of them in South Asia and more than a third in Sub-Saharan Africa. As a result, since March 2020 over 215 countries and territories have implemented 1,414 social protection measures to respond to the pandemic and ensuing economic crisis. 
Social assistance programs account for 62 percent of all social protection response measures, half of them being cash-based transfers of some sort.

This major shock has revealed the many challenges governments face when attempting to quickly respond to crises in order to protect the poor and vulnerable. Providing \textit{timely} assistance and support to those households most in need can increase their resilience and reduce the negative impacts of the shock on their short and medium-term well-being. Nonetheless, the lack of readily available and up-to-date socioeconomic data necessary to prioritize shock-responsive social protection measures is an important binding constraint for many governments in developing countries. 

This paper presents a portion of our work on a larger project with the World Bank to identify the most vulnerable populations in these countries. Having timely access to such information, particularly in data-deprived contexts, can improve the capacity of governments to design and operationalize better and more shock-responsive social protection measures. 

In collecting building data from OpenStreetMap, an open and crowdsourced platform described later, we noticed many untagged buildings with raw text data. We present a methodology to dynamically extract and structure the location of additional schools and health clinics in 10 West African and Sahelian countries. 

The remainder of the paper proceeds as follow: section two describes the OpenStreetMap (OSM) unstructured data as well as the pipeline set up to intake it. Section three describes the methodology used to structure the OSM data in order to classify it and identify the infrastructure acess points of interest, i.e. schools and clinics. Section four presents the results and preliminary validations of this methodology while section five concludes.

\section{Data}

\subsection{OpenStreetMap Data}

The OpenStreetMap project was founded in 2004 with the initial goal of mapping the U.K. as an alternative to the restricted government programs at the time. The OpenStreetMap Foundation (OSMF) was founded two years later to strengthen the advancements and prosperity of OSM. 

Originally, data was collected by volunteer mappers using public satellite images, GPS traces and ground surveys. Yahoo and Bing later authorized the use of their aerial photographs which contributed to the project's data entry system. Over the next several years, data began to be aggregated from multiple third-party sources. Subsequently, major development of the API enabled robust tracking of users and changes made to the data. After this improvement, the registered users catapulted from 100,000 to 1,000,000 users resulting in continual growth into the system in place today.

Now, OSM strives to be a free-to-use, open source database that covers the world entirely. Data is still primarily collected by thousands of contributors who perform ground surveys using various tools and amend their findings to the database. Additionally, OSM now uses numerous public government sources such as the Prototype Global Shoreline database from the NOAA, the European Union Corpernicus (GMES) database, et al. 

The OSM database consists of three data types: nodes, ways, and relations. Nodes represent a latitude and longitude of a physical location. Ways have two forms: open and closed. Open ways represent a line or boundary (e.g. rivers, road networks, walls, etc.). Closed ways represent polygons or areas, i.e., enclosed filled areas of territory. Relations are compositions of the previous two types consisting of an ordered list of nodes, ways and/or relations that define the logical relationships between elements on the map. This also allows for the creation of multipolygons, a relation of closed ways that allow for more complex shapes.

\begin{figure*}
   \includegraphics[scale=0.75]{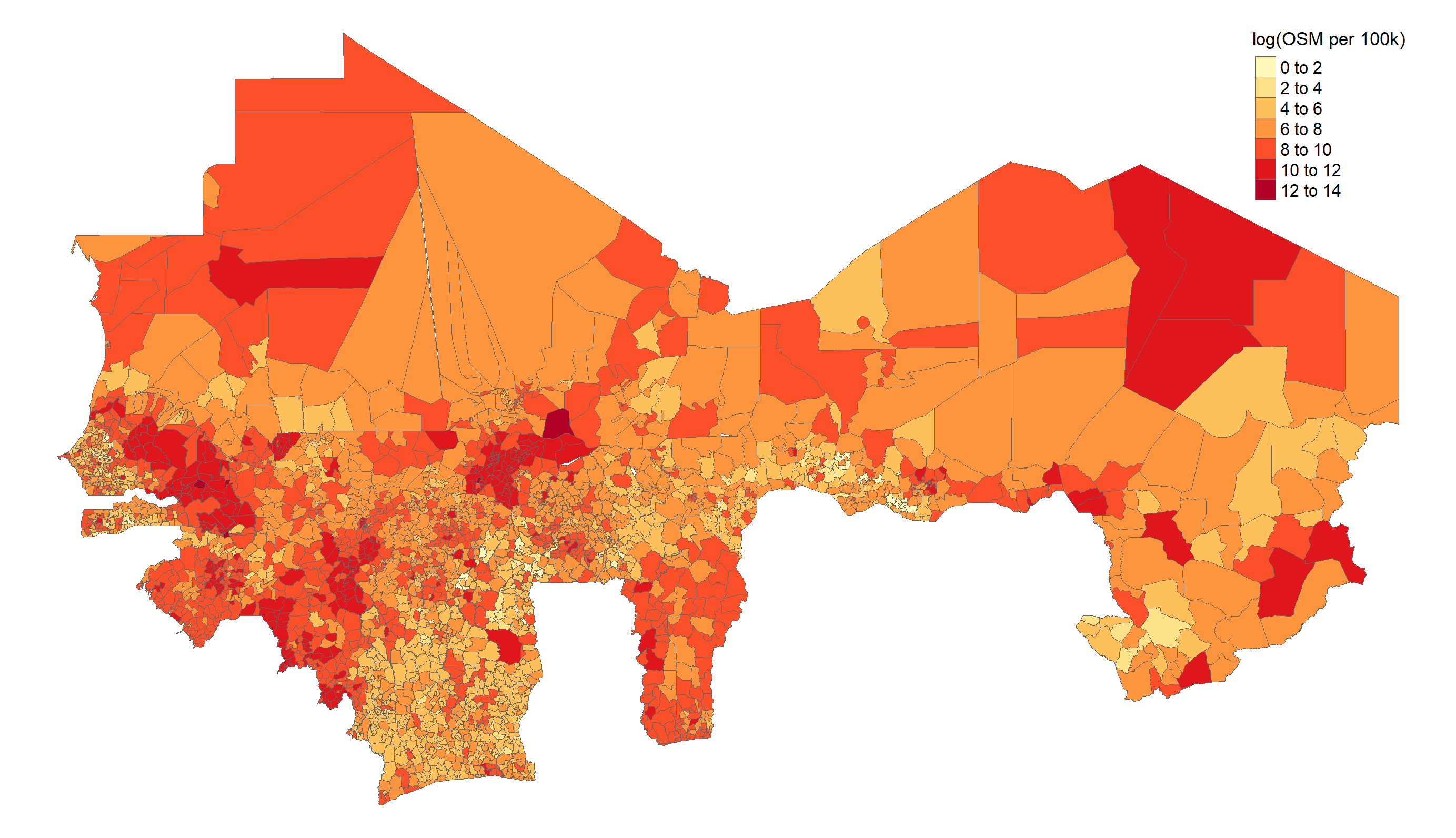}
   \caption{Map of the log of OSM data per 100,000 persons}
   \label{fig:OSM_per_100k}
 \end{figure*}

\begin{figure*}
   \includegraphics[scale=0.74]{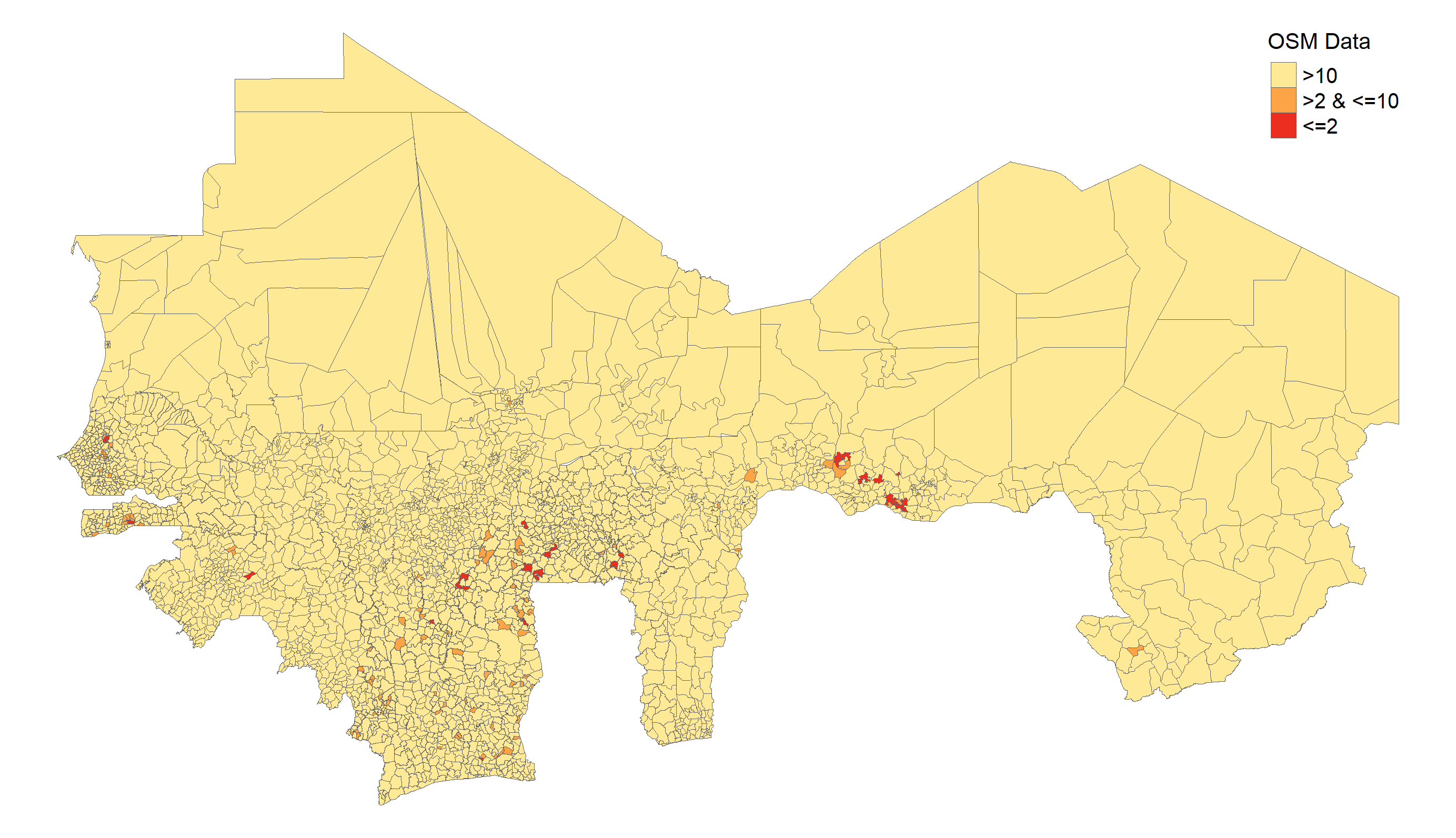}
   \caption{Map of areas missing OSM data ($\leq 2$ or $\leq 10$ data\ points)}
   \label{fig:OSM_missing}
 \end{figure*}

Figure \ref{fig:OSM_per_100k} presents the quantity of OSM data per 100,000 persons over the ten countries of our study in Sub-Saharan Africa\footnote{These countries are Burkina Faso, Niger, Guinea, Senegal, Cote D'ivoire, Mali, Togo, Mauritania, Benin, Chad and were chosen based on the need to identify school and clinics in these countries for another project.}. We see heterogeneity in the amount of OSM data per capita, with some locations having many more OSM data per capital than others. We are particularly concerned with complete gaps in OSM coverage, that is areas that have no or very little OSM data. Figure \ref{fig:OSM_missing} presents locations at the area 3 level having two or fewer (or ten or fewer) OSM data points overall. There are some areas that show no OSM data coverage whatsoever, and these seem to be limited to areas with lower population. We calculate that areas with missing OSM data have an average population of 28,000 persons. They also appear to exhibit geographic clustering.

The database structures these data using a tagging system for primary features but also includes a field to insert unstructured text. Each primary key represents a category of geographical elements such as buildings, highways, parks, etc. Keys that describe broad categories are assigned tags that label the subdivisions within the key. Furthermore, other properties are stored in metadata including: addresses, annotations, names, properties, references, and restrictions.

\subsection{Related Work}

The lack of reliable data from which to assess the vulnerability of populations in the developing world is a continual challenge \cite{serajuddin2015data}. Many researchers and organizations have theorized the potential that Big Data -- in particular unstructured data sources such as images, text, and video -- holds for meeting these data gaps \cite{mcbride2021predicting}. Several of these approaches use proprietary sources of data, such as Twitter, Google Maps, proprietary satellite imagery, or mobile phone metatdata \cite{kondmann2020combining, xie2016transfer, engstrom2017poverty, babenko2017poverty, blumenstock2015predicting}. Researchers have pointed to the risk of using closed, proprietary systems for development purposes \cite{hersh2020open}, leading to expanded interest in the use of open and freely available sources of Big Data \cite{perez2017poverty}. OpenStreetMap is one of those sources.

\clearpage

 \clearpage

OSM data has been used to fill data gaps in a variety of development contexts, such as using OSM road network data for predicting poverty \cite{zhao2016openstreetmap}, using OSM tags to estimate food security \cite{quinn2016openstreetmap}, as well as using OSM data to monitor progress in refugee settlements \cite{van2021development, feldmeyer2020using}. We differ from these approaches in that we focus on the suitability of OSM data to proxy for the location and availability of public services.

\begin{table}
\renewcommand{\thefootnote}{\fnsymbol{footnote}}
\captionsetup[table]{labelformat=empty,skip=1pt}
\begin{longtable}{lccccc}
\caption*{
\large Table 1: Comparison of Proportions of OSM Data
} \\ 
\toprule
Country & Schools & Clinics & Unresolved Tag & Other$^\thefootnote$ \\ 
\midrule
Benin & 0.19\% & 0.004\% & 91.681\% & 8.126\% \\ 
Burkina Faso & 0.272\% & 0.026\% & 76.071\% & 23.631\% \\ 
Cote d'Ivoire & 0.048\% & 0.054\% & 83.902\% & 15.996\% \\ 
Guinea & 0.069\% & 0.008\% & 66.288\% & 33.635\% \\ 
Mali & 0.02\% & 0.003\% & 88.89\% & 11.087\% \\ 
Mauritania & 0.022\% & 0.005\% & 65.193\% & 34.78\% \\ 
Niger & 0.013\% & 0.01\% & 85.451\% & 14.527\% \\ 
Senegal & 0.059\% & 0.01\% & 53.245\% & 46.686\% \\ 
Chad & 0.071\% & 0.013\% & 83.736\% & 16.179\% \\ 
Togo & 0.039\% & 0.011\% & 93.234\% & 6.715\% \\ 
\midrule
\textbf{Median} & \textbf{0.058\%} & \textbf{0.01\%} & \textbf{85.275\%} & \textbf{17.096\%} \\ 
\bottomrule
\end{longtable}
\vspace{2ex}
\footnotesize{$^{\thefootnote}$\emph{Other} refers to data points with keys/tags not related to schools or clinics. \emph{Unresolved Tag}  describe data points without a key/tag.}
\end{table}

\subsection{Data pipeline and raw data}

With the goal of querying comprehensive datasets of ten countries, we deviate from the traditional OSM pipeline. An entire country's geospatial data is too large for the API to distribute, therefore, we opt for a public data extract service, Geofabrik, which compiles large regions of OSM data into Protocolbuffer Binary Format, a format allowing for serialization of large data. Acquisition using this method allows for rapid downloads and mapping of the data. We then store the data on a server in the cloud for simple accesss and it enters the data cleaning process. Particularly observing schools and clinics, we apply a filter to the OSM data reducing to nodes and closed ways results in a lower dimensional dataset involving exclusively buildings and properties. These are later analyzed to identify schools and clinics specifically.

The raw OSM data once deserialized becomes a wide form table with rows being a node or closed way and columns being keys, tags, metadata, and the geometry. Since most locations are assigned one or two keys, much of the table contains empty values. The fashion in which the keys are input is fairly inconsistent. A number of tags that subset keys correlate with other keys. For example the keys "healthcare" and "building:healthcare" are both present but are treated differently nonetheless. This results in some rows having the values in one column while others store the same values in a separate column adding unecessary complexity to the table. Furthermore, a column labeled "tags" contains unstructured text containing various information such as keys and tags that were not input in a standard format. These unstructured tags contain business names, manual descriptions of the location, and any other information the mapper sees fit. For each location, the OSM database stores these in a string of characters resembling a JSON format. If a given data point has no specified key or tag, we refer to this data point as having an "unresolved tag" as it may still have information in the tags column.

Finally, another dimension to the raw data we have to overcome is the inclusion of multiple languages. Since many of the countries in West Africa primarily speak French, a proportion of the data is written in French along with some Arabic rather than English. This creates conflicts with encoding and the unsupervised learning performed on the data. Since many French words have similar roots to English, the model is still able to identify a meaningful amount of locations in French, but it must be stated that the model has a bias against non-English tags.

\subsection{Disparities in OSM data}

Table 1 presents evidence on the quality of OSM data in each country of study. For each country we calculate the average fraction of OSM data that contain (structured) keys for schools and clinics, as well as share of OSM data that contain unresolved tags as well as other tags or keys that presumably can be resolved to other types of buildings. The median country-level percentage of OSM data pertaining to schools, clinics, unresolved tags, and other keys is 0.06\%, 0.01\%, 85.3\%, and 17.1\% respectively.

Six countries hold 83\% or more in unresolved tags: Benin, Cote d'Ivoire, Mali, Niger, Chad, and Togo. Curiously, Benin has a high relative percentage of schools classified at 0.19\%, but very low share of clinics and other keys at 0.004\% and 8.13\% respectively. The median of the other five countries for schools classified is notably below the median of all countries at 0.03\%. Strikingly, Cote d'Ivoire has the highest percent of classified clinics, being the only country in the raw data having more clincs than schools at 0.05\% to 0.04\%.

Among the countries with below average percentage of unresolved tags, Burkina Faso has a significant amount of schools at 0.27\%. It has high percentages of clinics at 0.02\% and other keys at 23.6\%.



\section{Methodology}

\subsection{Data Wrangling}
With a proper data pipeline, data aquisition is somewhat elementary. A basic call of the Geofabrik database and deserialization is sufficient to begin scrutinizing the data. 

Aggregating the keys to a single discrete column of building types later used to filter down to schools and clinics. We accomplish this by using a vectorized function scanning through each column until the corresponding key is found. A number of keys have respective values that contain a more specific description of the building than the key itself. If a given key has such a value, it is prioritized over the key. If no key is found, we look at that record's "tags" column to mine any information that could improve our attempt to classify the building.

\begin{table*}[t!] 
\renewcommand{\thefootnote}{\fnsymbol{footnote}}
\captionsetup[table]{labelformat=empty,skip=1pt}

\setlength{\tabcolsep}{0.25em} 
\begin{longtable}{lccccccccc}
\caption*{
\large Table 2: Comparison of OpenStreetMap Data per 100,000 People\\ \small  } \\
\toprule
& & \multicolumn{4}{c}{Original (Keys)} & \multicolumn{4}{c}{Topic Modeled (Keys + Unstructured Tags)} \\ 
 \cmidrule(lr){3-6}\cmidrule(lr){7-10}
Country & Total & Schools & Clinics & Unresolved Tag & Other & Schools & Clinics & Unresolved Tag & Other \\ 
\midrule
Benin & 7455.39 & 14.13 & 0.28 & 7306.140 & 134.84 & 35.18 & 8.32 & 6822.56 & 589.33 \\ 
Burkina Faso & 4603.63 & 12.54 & 1.20 & 4512.940 & 76.95 & 33.22 & 5.73 & 3492.33 & 1072.35 \\ 
Cote d'Ivoire & 6134.35 & 2.96 & 3.34 & 5823.000 & 305.05 & 22.16 & 18.36 & 5133.84 & 959.99 \\ 
Guinea & 7262.19 & 4.98 & 0.59 & 5346.690 & 1909.93 & 15.00 & 5.88 & 4805.03 & 2436.28 \\ 
Mali & 22898.67 & 4.62 & 0.61 & 22173.040 & 720.40 & 113.94 & 14.03 & 20340.82 & 2429.88 \\ 
Mauritania & 888.61 & 0.19 & 0.04 & 790.400 & 97.98 & 3.74 & 1.35 & 568.91 & 314.61 \\ 
Niger & 20140.68 & 2.58 & 1.95 & 19766.320 & 369.83 & 44.42 & 16.91 & 17197.11 & 2882.24 \\ 
Senegal & 4247.86 & 2.52 & 0.44 & 2713.210 & 1531.69 & 17.39 & 8.12 & 2240.28 & 1982.07 \\ 
Chad & 70005.87 & 5.00 & 0.93 & 6824.410 & 63175.53 & 8.03 & 4.49 & 5856.08 & 64137.27 \\ 
Togo & 10978.69 & 4.33 & 1.24 & 10832.980 & 140.14 & 20.68 & 9.26 & 10219.84 & 728.91 \\ 
\midrule
\textbf{Median} & \textbf{7358.79} & \textbf{4.475} & \textbf{0.77} & \textbf{6323.705} & \textbf{337.44} & \textbf{21.42} & \textbf{8.22} & \textbf{5494.96} & \textbf{1527.21} \\ 
\bottomrule
\end{longtable}
\vspace{2ex}
\footnotesize{\emph{Other} refers to data points with keys/tags not related to schools or clinics. \emph{Unresolved Tag}  describe data points without a key/tag.}
\end{table*}

\subsection{Topic Modeling}

There are two standard methods of classification: supervised and unsupervised. A supervised approach involves using a source for validation holding either text or a form of word embeddings. In these sources resides keywords, keyphrases, descriptors, and indexing terms. After preprocessing the input, these supervised methods validate using the source vocabulary. This requries the input data to be a compilation of the validation source in order to be effective. There are a number of open source libraries that accomplish this in a refined approach.

\begin{figure}
\centering
   \includegraphics[width=25em]{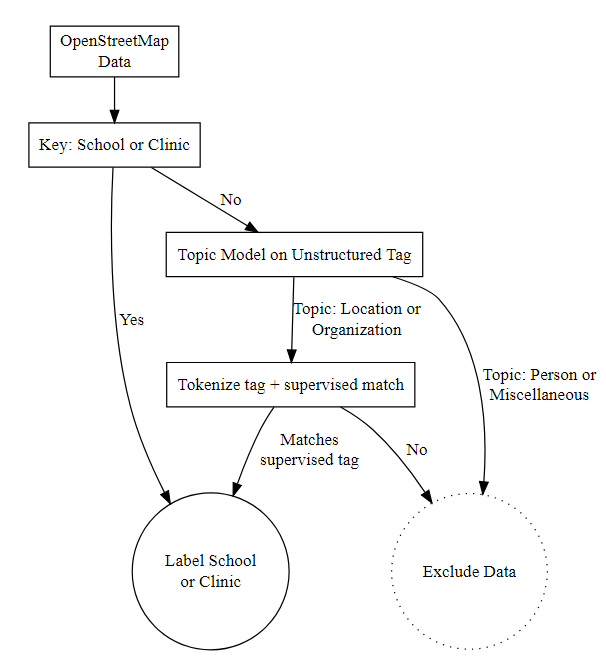}
   \caption{Flow diagram of topic model algorithm to classify schools and clinics from raw OSM data}
   \label{fig:topic_model_diagram}
 \end{figure}
 
An unsupervised approach involves a model containing probabilities of varying clusters of sequential words called "topics." When a model is trained, it develops topics based off of the context of word clusters. These topics are used to classify input text into a topic, clustering text together. Some models have pretrained topics with defined topic names to add more utility to the topics. Unsupervised models may be preferred in some use cases as they can take any input of text and attempt to classify it into a topic.

We summarize our algorithm in figure \ref{fig:topic_model_diagram}. Since the excess tags in the OSM data may contain keys we have already observed, we opt for both methods to maximize the possible information we can extract from the limited data we have in these ten countries. For the supervised portion, we already have a predefined vocabulary in the OSM keys. After removing punctuation and other unnecessary characters, we cross-check the text with the OSM keys to identify if there is a match. This reinforces our efforts to limit the number of discrete values of the data. For our objective, we limit this to any value that falls under schools or health clinics. Some of these values include: primary and secondary schools, research institutes, hospitals, first aid facilities, etc. 

In cases where no key indicating a OSM building is a school or clinic, the information indicating whether it is in fact a school or clinic may be contained within the unstructured tag. For these cases, we implement a unsupervised model, MIT Information Extraction (MITIE), to look through the unstructured text describing the building and to classify these as pertaining to schools or clinics. MITIE offers a tool to train topics for a model, but we opt to use a pretrained model consisting of four topics: person, location, organization, and miscellaneous. The model is trained on English words, but successfully classified many French words as well. If the model classifies any text as a location or organization, we utilize the values to fill in remaining gaps in the data when possible. Any location or organization found by the model whose value indicates a school or clinic are later replaced with an appropriate OSM key consistent with the rest of the data.

For example, consider the following data: "\{`type` :`boundary`, `place` : `language school`\}" and "\{`source date` : `21/03/2021`, `name` : `Niger hospital`\}." The first string contains the OSM key "school" so it is immediately classified as a school. The second string, however, does not contain an OSM key. The MITIE model will break the string into tokens as "source date", "21/03/2021", "name", and "Niger hospital." The model then classifies the tokens into one of the four topics. "Niger hospital" would get classified as location, and we group hospitals under the OSM key "clinic." The end result of the data is classifications as school and clinic respectively.

\begin{figure}
    \centering
   \includegraphics[scale=0.75]{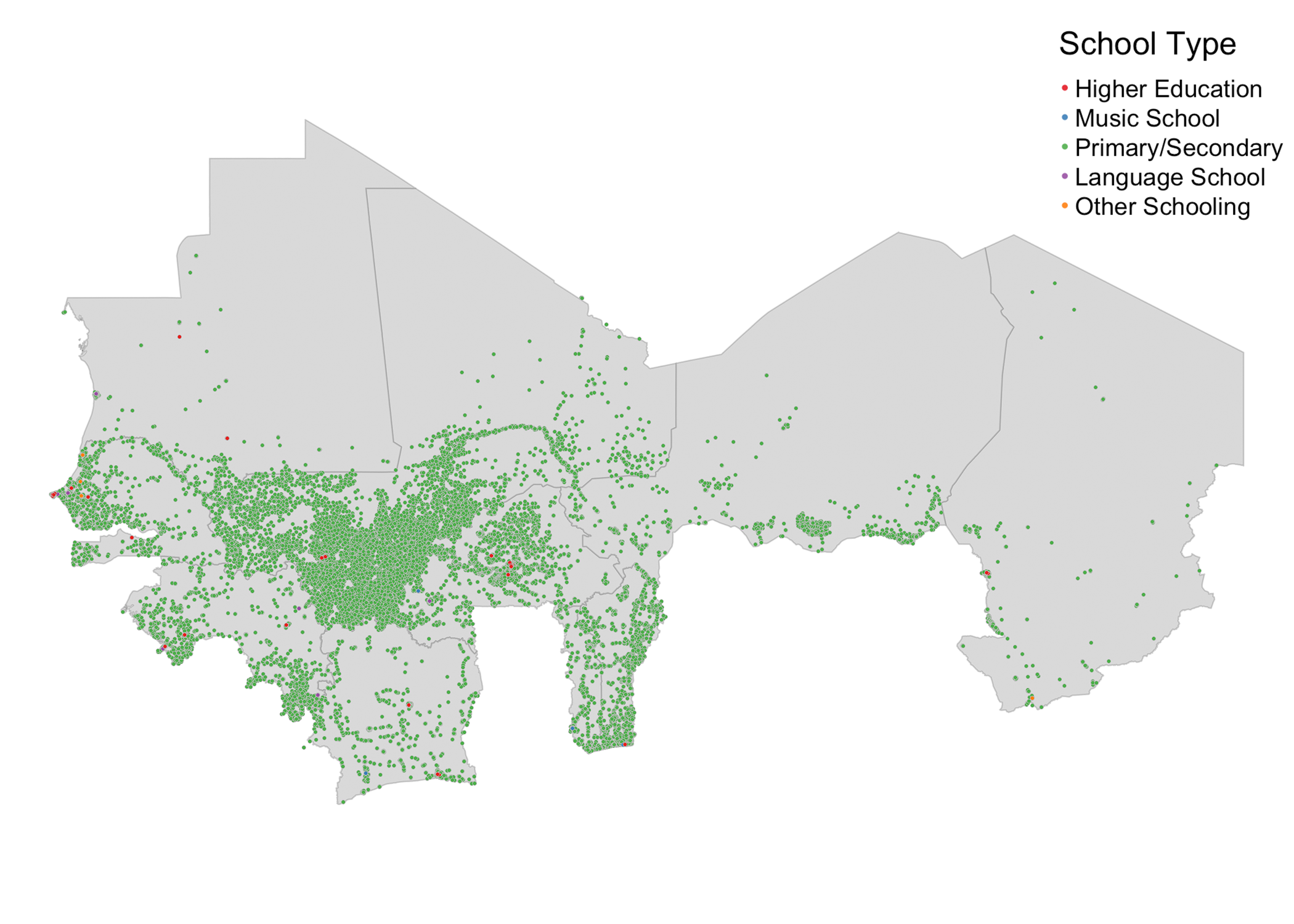}
   \caption{OpenStreetMap data on schools shown in ten West African countries}
   \label{fig:map_school_final}
 \end{figure}
 \begin{figure}
 \centering
   \includegraphics[scale=0.62]{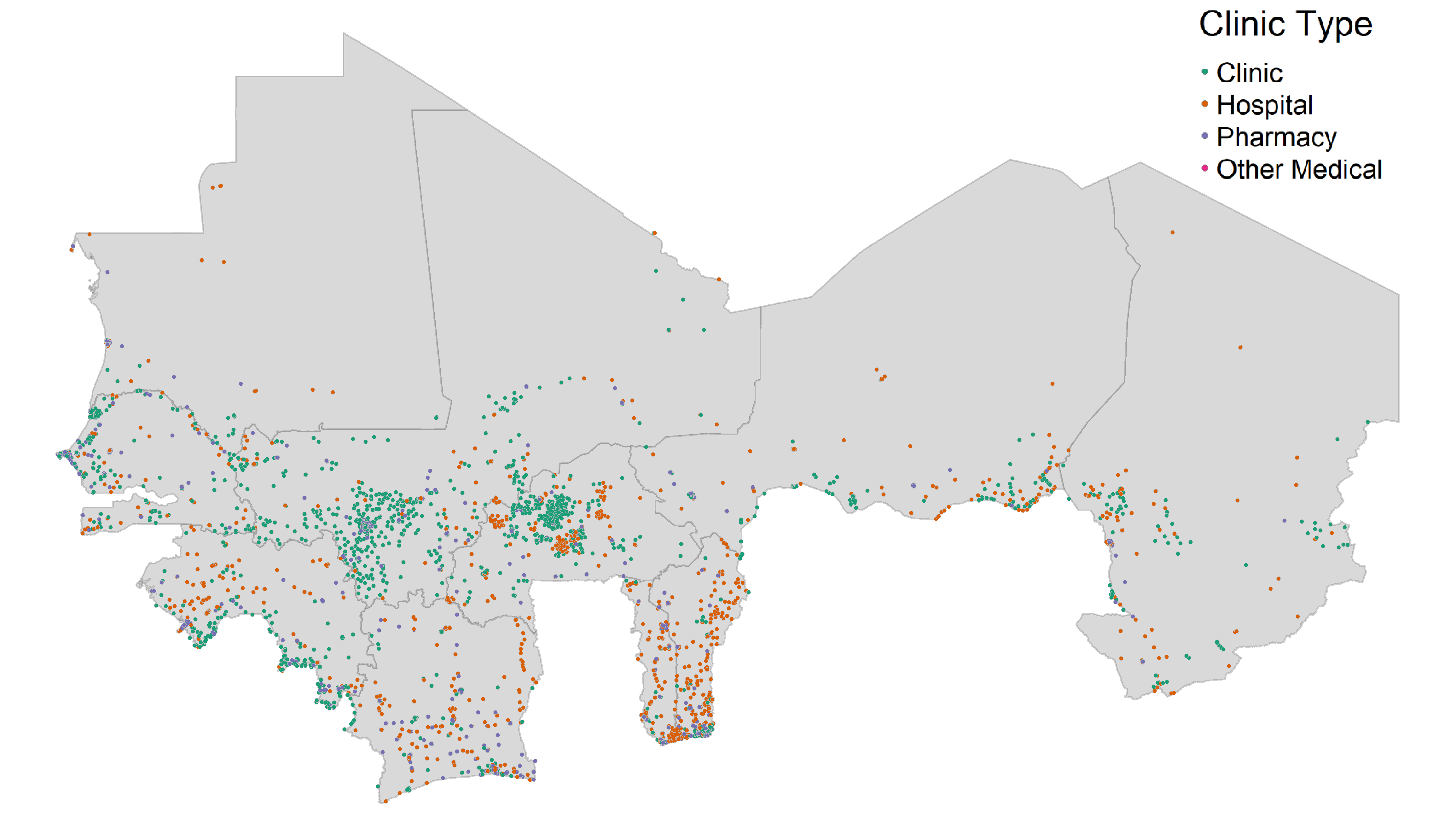}
   \caption{OpenStreetMap data on clinics shown in ten West African countries}
   \label{fig:map_clinics_final}
 \end{figure}
 \clearpage

\section{Results and Validation}

\begin{figure}
    \centering
   \includegraphics[width=25em]{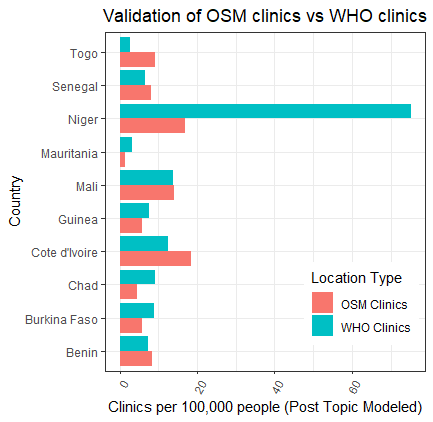}
   \caption{Comparison of data sources of clinics in West Africa}
   \label{fig:clinics_valdation}
 \end{figure}

\subsection{Results}
Considering the amount of unresolved tags in the original data, we should note the number of schools and clinics that may be unlabeled is considerably high. Because of this, we observe the density of data by population in order to understand where data is being generated and where it may be missing. Table 2 presents OpenStreetMap data per 100,000 people before and after the topic model cleaning approach. We find that the topic model is able to substantially increase the number of clinics and schools identified. Roughly twice to fifty times as many schools and clinics are identified after the cleaning step. 

We find substantial imbalances in the schools and clinics per capita identified in the OSM data. Mali and Niger show a large number of schools identified, while Mauritania, Chad, Guinea and Senegal show comparitively few. We see a larger number of clinics per capita in Cote d'Ivoire, Mali, and Niger, with fewer in Mauritania, Chad, and Guinea. 

Mali, Mauritania, Niger, and Chad all have quite high variance in the density of data. Mali and Niger while having large amounts of data, have a low amount of data that can be tagged. Mauritania have much less data per 100,000 people overall while also having very few classified values. In contrast, Mali has a high population with a high number of classified schools relative to the other countries despite its numerous unresolved tags.

Looking at schools, we aggregate locations into five different categories: Higher Education, Primary/Secondary, Music, Language, and Other. Figure \ref{fig:map_school_final} shows the distribution of schools we are able to classify among the ten countries. Mali has noticably more schools tagged than any other country. This is consistent with Mali being one of the three countries with significantly more data.

Chad and Mauritania appear to have a significant amount of schools absent of labels. These two countries in particular are noteworthy to have this issue as they have the largest and least amount of data per 100,000 people respectively. Mauritania is a much less populated country than the rest and is expected to have less density of buildings although it still has 64\% of its data unclassified. 


In regards to clinics, the raw numbers show the mass amount of classified clinics in Burkina Faso and Mali. On the otherhand, we see some deviation from the trends we see in schools. Hospitals as well as other medical facilities are reasonably distributed between most countries with the exception of some clusters in Burkina Faso and Togo seen in Figure \ref{fig:map_clinics_final}. 

This reveals disparities in the density ratio of data in some contries. Mali, Benin, Burkina Faso, Guinea, and Togo all have notable density of schools, but much less density of clinics when population is taken into account. Meanwhile, Cote d'Ivoire has a higher density of clinics than schools relative to the other countries. 
The density of clinics in most countries is inconsistent with its respective density of schools in reguards to population.


\subsection{Validation}

To assess the validity of the schools and clinics identified by our method using the OSM data, we validate our findings against alternative sources. We opt to compare the frequency of schools and clinics within each country to scrutinize the trends in the OSM data to note whether it is inconsistent in another source, aiding our effort to see what information may be missing from the database.

Our validation source for clinics comes from the World Health Organization (WHO) \cite{maina2019spatial}. It contains a vast list of health facilities aggregated from a variety of sources. The data excludes private and specialized facilities. The sources are partially government and non-government based. The non-government sources include Geonames, Microsoft Encarta, or a combination of several others. Last updated in early 2019, it holds information on 100,000 public health facilities in sub-Saharan Africa.

There were a number of countries where OSM contained more clinics than the WHO database. In Figure \ref{fig:clinics_valdation}, we see Benin, Cote d'Ivoire, Mali, Senegal, and Togo all hold the same or more health facilities in OSM as in the WHO database. Cote d'Ivoire and Togo specifically heavily outperform in OSM data, suggesting most of their unresolved tags describe buildings other than clinics.

Countries that underperform in OSM clinics include: Burkina Faso, Chad, Guinea, Mauritania, and Niger. Niger in particular has an eye-catching difference in frequency. The WHO database suggests than many of the unresolved tags in Niger describe clinics. It should be taken into consideration that all countries have significant unresolved tags that likely would increase the known frequency of clinics considerably given enough clinic locations with a given country's unresolved tags.

Unfortunately, we have been unable to identify a quality validation set for schools. We are unaware of any quality dataset that has a significant amount of data on schools in our ten countries. We are open to any references to alternative data sources to OSM on schools in sub-Saharan Africa.

\section{Conclusions}

We find that our topic modeling approach improves substantantially the amount of schools and clinics identified in the raw OSM data. Between two to fifty times as many schools and clinics are identified after parsing through the unstructured text using our topic modeling approach. This presents a promising path forward for researchers looking to use this data for the purposes of understanding the location of public goods. 

While subnational validation of this methodology to extract, structure and intepret OSM data remains to be investigated, it is clear that the quantity and quality of data in the countries selected is not uniform. Quality seems to be even more heterogeneous across countries than quantity, as well as more deficient in general. Nonetheless, Chad is a clear outlier both in terms of quantity and quality of data, while Senegal stands out in data quality relative to other countries.

The underlying factors driving the heterogeneity of OSM data in the region are unclear, however the natural data production processes leave many areas deprived of enough and, more importantly, good enough data.  Even after being processed, over 70\% of the OSM data points extracted are not classifiable and thus can't be exploited for analytic or policy purposes. 

In order to improve the quantity and quality of OSM data in the region, a public investment in the production of exploitable OSM data could be devised. Data should be considered an essential input for the economy of the future, much like labor and capital have been so far. Thus, such an investment is a public good that could have multiple spill overs for policy and private sector purposes. 
Investing in the production of exploitable OSM data in the region could be done through a cash for work program focused on digital data collection. Such a program would bootstrap relevant digital skills into the labor force of the  countries considered and help to slow or even reverse the digital divide and digital inequalities faced by much of the developing world.

\section{Acknowledgments}

The authors gratefully acknowledge support by the Kay Family Foundation. Elmer Camargo provided excellent research assistance. 

\bibliographystyle{unsrt}  
\bibliography{Topics} 

\appendix

\section{Research Methods}

Analysis was conducted in R and Python. 

\section{Online Resources}

Reproduction code is available at \url{https://github.com/HershLab/vulnerability-mapping-data-prep}
\end{document}